\documentclass[runningheads]{llncs}
\usepackage{graphicx}
\usepackage{tabularx}
\usepackage{enumitem}
\usepackage{amsmath}
\usepackage{amsfonts}
\usepackage{hyperref}
\begin{document}
\title{DefenseVGAE: Defending against Adversarial Attacks on Graph Data via a Variational Graph Autoencoder}
\titlerunning{DefenseVGAE}
%
%
\author{Ao Zhang
 \and
 Jinwen Ma}
%
\authorrunning{A. Zhang et al.}
%
\institute{Department of Information Science, School of Mathematical Sciences, Peking University, Beijing, 100871, China \\ \email{zhangao520@pku.edu.cn}, \quad \email{jwma@math.pku.edu.cn}
}
\maketitle              
\begin{abstract}
Graph neural networks (GNNs) achieve remarkable performance for tasks on graph data. However, recent works show they are  extremely vulnerable to adversarial structural perturbations, making their outcomes unreliable.  In this paper, we propose DefenseVGAE, a novel framework leveraging variational graph autoencoders(VGAEs) to defend GNNs against such attacks.  DefenseVGAE is trained to reconstruct graph structure. The reconstructed adjacency matrix can reduce the effects of adversarial perturbations and boost the performance of GCNs when facing adversarial attacks. Our experiments on a number of datasets show the effectiveness of the proposed method under various threat models. Under some settings it outperforms existing defense strategies. Our code has been made publicly available at \url{https://github.com/zhangao520/defense-vgae}. 

\end{abstract}
\section{Introduction}
Graphs are a natural representation that can model diverse data from nearly all scientific and engineering applications, such as social networks, citation networks, molecular structure. Not surprisingly, machine learning on graph data has a longstanding history. Specifically, graph neural networks (GNNs) starts to push forward the performance of several fundamental tasks ranging from node classification\cite{Kipf2016SemiSupervisedCW}, over community detection\cite{Zhang2018SINESI}, to generative modeling\cite{Kipf2016VariationalGA}.

Here we focus on semi-supervised node classification – given a single large (attributed) graph and the class labels of a few nodes the goal is to predict the labels of the remaining unlabelled nodes. Graph Convolutional Networks (GCNs)\cite{Kipf2016SemiSupervisedCW} and their recent improvements\cite{Li2018AdaptiveGC,Zhuang2018DualGC} have shown great success in node classification tasks by performancing convolution operations aggregating and combining the information of neighbor nodes in the graph domain.

Despite GCNs have improved the state of the art in node classification,  they can be attacked on multiple front - changing only a few links or node attributes which are unnoticeable to the users can lead to completely wrong predictions\cite{Zgner2018AdversarialAO,Zugner2019AdversarialAO,Wang2019AttackingGC,Dai2018AdversarialAO}. Two types of threat models have been considered in recent literatures: under the targeted attack model, the attacker aims to let the trained model misclassify some target nodes; whereas in the untargeted attack model, the attacker aims to hinder overall model performance on all test data.

This poor adversarial robustness is bound to arouse concerns about applying GCNs to real world  applications, especially those safety-critical domains such as fraud detection\cite{Wang2019FdGarsFD}, protein interface prediction\cite{Fout2017ProteinIP}, breast cancer classification\cite{Rhee2018HybridAO}.
Therefore, adversarial defence studies play a crucial role in graph domain and their goal is to detect or defend attacks and introduce strategies that are more robust against adversarial perturbations.

Various defenses have been proposed to mitigate the effect of adversarial attacks on graph data. 
There are three categories of defense mechanisms\cite{Jin2020AdversarialAA}:      (1)   changing GNN model’s manner of learning  to make the classifier more robust against attacks, e.g., graph adversarial training which injects adversarial examples into the training set\cite{Xu2019TopologyAA,Feng2019GraphAT}. (2)  attempting to purify the perturbed graph data\cite{Entezari2020AllYN,Wu2019AdversarialEF}. (3)  learning an attention mechanism to distinguish
adversarial edges and nodes from the clean ones\cite{Zhu2019RobustGC}

A few pieces of work\cite{Meng2017MagNetAT,Samangouei2018DefenseGANPC,Song2017PixelDefendLG,Hwang2019PuVAEAV,Ghosh2018ResistingAA} in the existing literature on defense against adversarial attacks on image domain have attempted to use generative models to purify input data that may have added adversarial perturbations. These methods mostly work by using a generative model to learn the data distribution and projecting the adversarial examples to the manifold of legitimate, or natural examples. Inspired by this methodology, we propose a novel defense mechanism which is effective against adversarial attacks on graph data. We propose to leverage a variational graph autoencoder(VGAE) to encode a graph with adversarial perturbed structure into a latent representation matching a prior distribution and reconstruct graph structure from the latent representation, then we recover the reconstructed adjacency matrix's sparsity and use it to train a GCN model which can significantly resist the structure attacks.

In summary, our contributions are as follows:
\begin{itemize}

    \item We propose a VGAE-based defense method, which we call DefenseVGAE to effectively resist against adversarial attacks on graph data. The proposed method   has a desirable effect of suppressing the adversarial perturbations.
    \item Experimental results demonstrate the effectiveness of our proposed method on on three benchmark graph datasets under different threat models.
\end{itemize}

The rest of the paper is organized as follows.  In Section 2, we review necessary background regarding GCNs, known attack models, defense mechanisms and VGAEs. Our proposed method is formally motivated and introduced in Section 3. Experimental results, under different threat models, as well as comparisons to other defenses are presented in Section 4. Finally, we conclude the paper in Section 5.

\section{Related Work}
In this work, we propose to use VGAEs for the purpose of defending against adversarial attacks in semi-supervised node classification task. Before specifying our approach in the next section, we explain related work in four parts. First, we describe the graph convolutional network paradigm as representative for victim models. Next,
we discuss different attack methods employed in the literature. We, then, go over related defense mechanisms against these attacks and discuss their strengths and shortcomings. Lastly, we explain necessary background information regarding VGAEs.

\subsection{Graph Convolutional Network}

Given an undirected, unweighted graph $\mathcal{G}=(\mathcal{V}, \mathcal{E})$ with $N$ nodes $v_{i} \in \mathcal{V}$, edges $(v_{i}, v_{j}) \in  \mathcal{E}$, an adjacency matrix $A \in \mathbb{R}^{N\times N}$ ,a degree matrix $D_{ii}=\sum_{j}A_{ij}$, and node features $X\in \mathbb{R}^{N\times D}$. A multi-layer GCN forward model then takes the simple form:

\begin{equation}
    H^{(l+1)}=\sigma\left(\tilde{D}^{-\frac{1}{2}} \tilde{A} \tilde{D}^{-\frac{1}{2}} H^{(l)} W^{(l)}\right)
\end{equation}

\noindent where $\tilde{A} = A + I_{N}$ and $\tilde{D}_{i i}=\sum_{j} \tilde{A}_{i j}$, and $\sigma$ is the activation function to introduce non-linearity, $H^{(0)} = X$, $W^{(l)} (l = 0, 1,\cdots,L-1)$ 
are weight matrices. A fully connected layer with softmax function is usually applied after $L$ layers of graph convolution layers for the classification.
A two-layer GCN is commonly considered for semi-supervised node classification tasks. The model can, therefore, be described as:

\begin{equation}
    Z=f(X, A)=\operatorname{softmax}\left(\hat{A} \operatorname{ReLU}\left(\hat{A} X W^{(0)}\right) W^{(1)}\right)
\end{equation}

\noindent where $\hat{A}=\tilde{D}^{-\frac{1}{2}} \tilde{A} \tilde{D}^{-\frac{1}{2}}$ is the symmetrically normalized adjacency matrix.

\subsection{Graph Adversarial Attacks}

The vulnerability of deep neural networks to adversarial attacks has generated a lot of interest and concern in the past few years\cite{Goodfellow2014ExplainingAH,Liu2017DelvingIT}.  Adversaries can manipulate deep-learning outputs by adding imperceptible perturbations on benign data. Most of the researches on adversarial machine learning are focused on algorithms to fool deep neural networks, mainly for the task of image classification.

Recently, some adversarial attack methods on graph data have been proposed to reveal the vulnerability of GCNs. Attack algorithms can be categorized into different types based on different goals, resources, knowledge and capacity of attackers\cite{Jin2020AdversarialAA}.

\begin{itemize}
    \item{ \textbf{Poisoning} or \textbf{Evasion}.} Evasion attacks happen after the GNN model is trained or in the test phase, while poisoning attacks happen before the GNN model is trained.
    \item {\textbf{Targeted} or \textbf{Non-targeted}.}
    Under the targeted attack model, the attacker aims to let the trained model misclassify some target nodes; whereas in the untargeted attack model, the attacker aims to hinder overall model performance on all test data.
    \item {Modifying \textbf{Feature} or \textbf{Edges}.}
    The attacker can insert adversarial perturbations from different aspects. The perturbations can be categorized as modifying node features or adding/deleting edges under certain budget of total actions.

\end{itemize}

\subsection{Graph Adversarial Defenses}

Various defense mechanisms have been employed to combat the threat from adversarial attacks. In what follows, we describe some representative defense strategies working on poisoning attacks.

\subsubsection{GCN-Jaccard}

Wu et al.\cite{Wu2019AdversarialEF} proposed a defense method by removing the edges whose two end nodes have small Jaccard similarity basing on an empirical observations of the attack methods: attackers tend to add edges which connect to nodes with different features. The experimental results demonstrate the effectiveness and efficiency of the proposed defense method, however this method can only work when the node features are available.

\subsubsection{GCN-SVD}

Entezari et al.\cite{Entezari2020AllYN}  observed that Nettack\cite{Zgner2018AdversarialAO}  results in changes in high-rank spectrum of the graph, which corresponds to low singular values. Thus they proposed to purify the perturbed adjacency matrix by using truncated SVD to get its low-rank approximation. However, our experiments show that GCN-SVD does not always guarantee a good performance on clean data.

\subsubsection{RGCN}

RGCN\cite{Zhu2019RobustGC}  enhances the robustness of GCNs against adversarial attacks by using Gaussian distributions in hidden layers and learning variance-based attention weights in aggregating node neighborhoods. Actually, the Gaussian distributions in hidden layer are also used in variational graph autoencoders, which are the basis of our approach.

\subsection{Variational Graph Autoencoder}

The variational graph autoencoder\cite{Kipf2016VariationalGA}
is a framework combining graph convolutional networks and variational inference which maps nodes into a latent feature space and decode graph information from from normal distribution.  VGAEs can be used to learn network embeddings or generate new graphs. 
For the variational graph encoder, we optimize the variational lower bound as follows:

\begin{equation}
    \mathcal{L}=\mathbb{E}_{q(Z |(X, A))}[\log p(\hat{A} | Z)]- \text{KL} [q(Z | X, A) \| p(Z)]
\end{equation}

\noindent Here, $\text{KL}[q(\cdot) \| p(\cdot)]$  is the Kullback-Leibler divergence between $q(\cdot)$ and $p(\cdot)$. $p(Z) = \prod_{i} p\left(\mathbf{z}_{i}\right)=\prod_{i} \mathcal{N}\left(\mathbf{z}_{i} | 0, I\right)$ is  a Gaussian prior.

\section{Methodology}
\begin{figure}[ht]
    \centering
    \includegraphics[width=\linewidth]{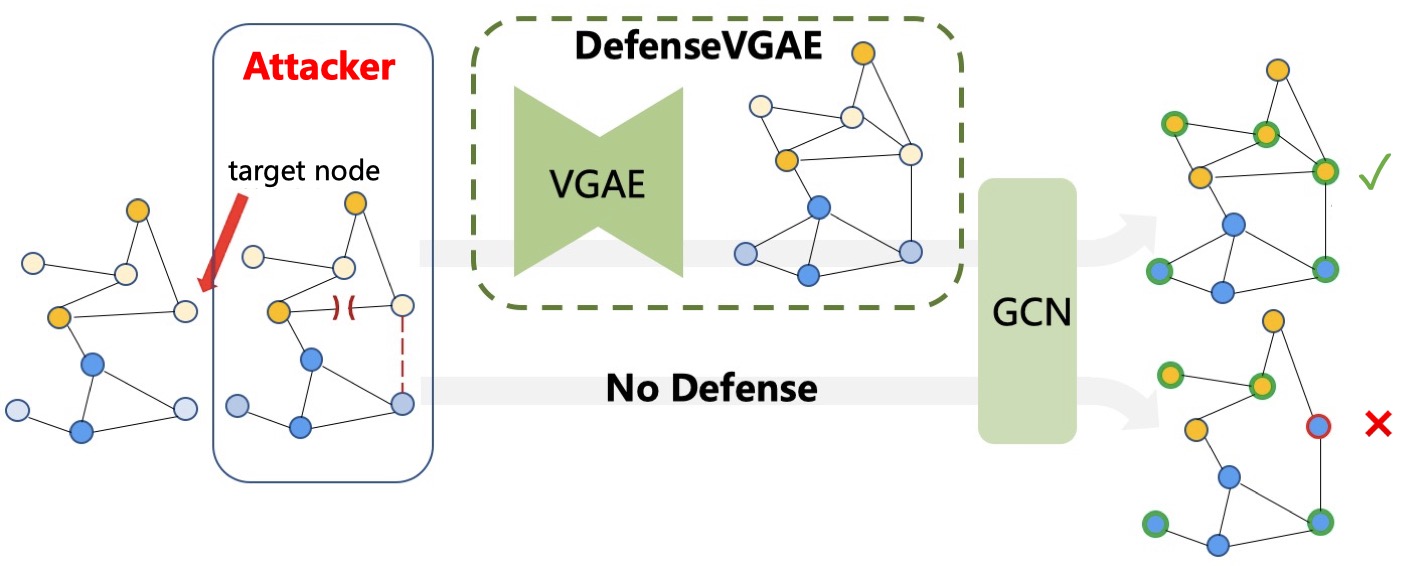}
    \caption{Overview of the DefenseVGAE algorithm.}
    \label{defensevgae-overview}
\end{figure}

We propose a new defense strategy which uses a VGAE trained on the perturbed graph data to reconstruct the adjacency matrix to mitigate adversarial perturbations. Given a perturbed graph $\hat{G} = (\hat{A}, \hat{X})$ which was attacked by adding  or removing edges  with regard to the clean graph data $ G = (A, X)$ - here we focus on graph structure attacks, i.e., $\hat{X} = X$. If we are not aware of this attack, the trained GCN model on $\hat{G}$ will misclassify the target nodes or have bad overall performance on all test data which could raise security concerns. In order to to mitigate malicious perturbations' impact, we trained a VGAE $g_{\hat{G}}$ on the graph $\hat{G}$ and reconstructed the adjacency matrix $\hat{A}$. When the reconstructed adjacency matrix $\tilde{A} = g_{\hat{G}}(\hat{A},\hat{X})$ is used, the GCN model can significantly resist the attacks. The overview of the proposed method is described in Figure \ref{defensevgae-overview}.

\subsection{Training Process of DefenseVGAE}

DefenseVGAE is comprised of a graph convolutional network (GCN) encoder and a simple inner product decoder. The encoder is defined by an inference model parameterized by a two-layer GCN:
\begin{equation}
    q(Z | A, X)=\prod_{i=1}^{N} q\left(\mathbf{z}_{i} | A, X\right)
\end{equation}

\begin{equation}
 q\left(\mathbf{z}_{i} | A, X\right)=\mathcal{N}\left(\mathbf{z}_{i} | \boldsymbol{\mu}_{i}, \operatorname{diag}\left(\boldsymbol{\sigma}_{i}^{2}\right)\right)
\end{equation}

\noindent Here $\boldsymbol{\mu}=\mathrm{GCN}_{\boldsymbol{\mu}}(A, X)$  is the matrix of mean vectors $\boldsymbol{\mu}_{i}$; similarly $\log \boldsymbol{\sigma}=\operatorname{GCN}_{\boldsymbol{\sigma}}(A, X)$. The two-layer GCN is defined as $\operatorname{GCN}(A, X)=\tilde{A} \operatorname{ReLU}\left(\tilde{A} X W_{0}\right) W_{1}$, with weight matrices $W_{i}$. $\mathrm{GCN}_{\boldsymbol{\mu}}(A, X)$ and $\mathrm{GCN}_{\boldsymbol{\sigma}}(A, X)$ share first-layer parameters $W_{0}$. Our decoder model is used to reconstruct structure $A$, predicting whether there is a link between two nodes by an inner product between latent variables.
\begin{equation}
    p(A | Z)=\prod_{i=1}^{N} \Pi_{j=1}^{N} p\left(A_{i j} | \mathbf{z}_{i}, \mathbf{z}_{j}\right), \text { with }  p\left(A_{i j}=1 | \mathbf{z}_{i}, \mathbf{z}_{j}\right)=\text{sigmoid}\left(\mathbf{z}_{i}^{\top} \mathbf{z}_{j}\right)
\end{equation}

\noindent We minimize the loss function as:

\begin{equation}
    \mathcal{L} = \mathcal{L}_{\text{RC}} + \mathcal{L}_{\text{KL}}
\end{equation}

\noindent Here

\begin{equation}
    \mathcal{L}_{\text{RC}} = \lambda \sum A_{ij} \log \hat{A}_{ij}+ \sum (1-A_{ij}) \log (1-\hat{A}_{ij}) 
\end{equation}

\begin{equation}
    \mathcal{L}_{\mathrm{KL}}=\boldsymbol{\mu}^{2}+\boldsymbol{\sigma}^{2}-\log \left(\boldsymbol{\sigma}^{2}-1\right) 
\end{equation}

\noindent As $A$ is very sparse, it can be beneficial to re-weight terms with $A_{ij} = 1$ in $ \mathcal{L}_{\text{RC}}$.

\subsection{Recover the Reconstructed Stucture's Spasity}

We denote $R_{A}$ as the proportion of the non-zero elements of $A$. $g_{G}$ is the VGAE trained on the given graph $G = (A,X)$. Then we have the reconstructed adjacency matrix
\begin{equation}
    \tilde{A} = g_{G}(A, X) 
\end{equation}

\noindent Next we make  $\tilde{A}$ a sparse matrix $\tilde{A^{*}}$ with a proper $R_{\tilde{A^{*}}}$ by the following principle:

\begin{equation}
\tilde{A^{*}}_{ij}=\left\{\begin{array}{ll}1 & \text { if } \tilde{A}_{ij}>p \\ 0 & \text { otherwise }\end{array}\right. \text{with }  p = \text{percentile}(\tilde{A}, 100(1-R_{\tilde{A^{*}}}))
\end{equation}

The work \cite{Zgner2018AdversarialAO} showed high degree nodes are slightly harder to attack: they have both, higher classification accuracy in the clean graph and in the attacked graph. Therefore we can perform a hyperparameter search on $R_{\tilde{A^{*}}}$ above $R_{A}$, i.e. we prefer a $\tilde{A^{*}}$ which is less sparse than $A$, and select the one with the best validation accuracy.

\section{Experiments}
In this section, we empirically evaluate the effectiveness of our proposed defense technique under different settings. 
\subsection{Experimental Settings}


\textbf{Dataset:} We use widely used Cora, Citeseer\cite{Sen2008CollectiveCI} and Polblogs\cite{polblogs} datasets. Dataset statistics are summarized in Table \ref{table:datasetstat}.
We closely follow the experimental settings in previous works\cite{Kipf2016VariationalGA,Entezari2020AllYN}.

Specifically, we adopt the same dataset splits as in \cite{Kipf2016VariationalGA} for Cora and Citeseer datasets, and for Polblogs we split the network in labeled (20\%) and unlabeled nodes(80\%). Half of the labeled data is used for training and the other half is used for validation in the process of training the GCN model.  We first train the GCN surrogate model on the labeled data and then we select 500 target nodes classified correctly from test set.

As Polblogs does not provide node features, the attribute matrix is set to an identity matrix by default.

\begin{table}[h]
    \caption{Dataset statistics}
    \centering
    \begin{tabular}{p{2.5cm}<{\centering}p{1.5cm}<{\centering}p{1.5cm}<{\centering}p{1.5cm}<{\centering}p{1.5cm}<{\centering}}
        \hline
        Datasets & Nodes & Edges  & Classes & Features \\
        \hline
        Cora      & 2708   & 5429   & 7     & 1433 \\
        Citeseer  & 3327   & 4732   & 6     & 3703  \\
        Polblogs  & 1490   & 19025  & 2     & 1490  \\
        \hline       
    \end{tabular}
    \label{table:datasetstat}
\end{table}

\noindent\textbf{Baselines and Adversarial Attack Methods:} To evaluate the performance of DefenseVGAE, we compare it with the natural trained GCN and three effective defense methods: GCN-Jaccard\cite{Wu2019AdversarialEF}, GCN-SVD\cite{Entezari2020AllYN} and RGCN\cite{Zhu2019RobustGC} under the targeted Nettack\cite{Zgner2018AdversarialAO} and untargetd Metattack\cite{Zugner2019AdversarialAO}. For both attacking methods, we focus on changing graph structures.

\noindent\textbf{Parameter Settings:}  We train VGAEs for 250 iterations for the Cora and Citeseer datasets, and 500 iterations for the Polblogs dataset using Adam\cite{Kingma2014AdamAM} with a learning rate of 0.001. We construct encoders with a 32-neuron hidden layer and a 16-neuron embedding layer for all the experiments. For GCN and baselines, we retain to the settings described in the corresponding papers. 

\subsection{Results on Clean Datasets}

\begin{figure}[ht]
    \centering
    \includegraphics[width=\linewidth]{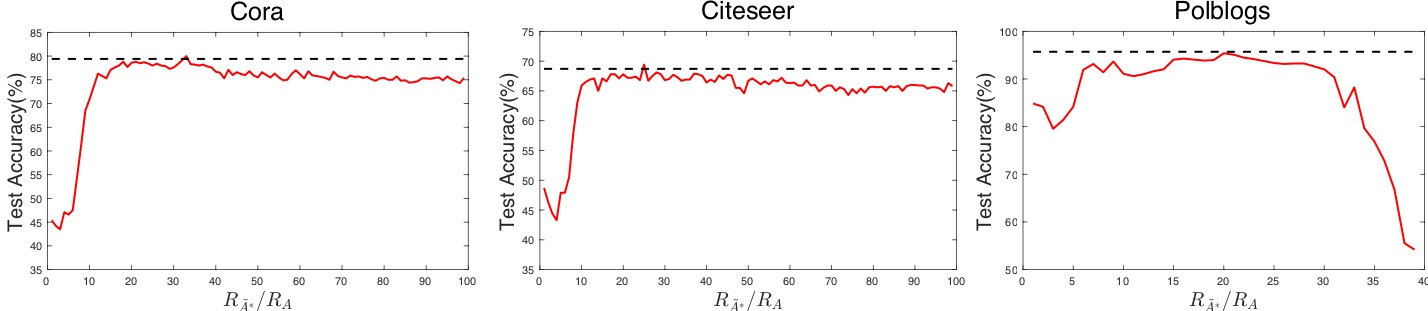}
    \caption{Red lines show
    test accurary under different reconstructed matrix's sparsity. Black dashed lines are  test accurary of the natural trained GCN models.}
    \label{fig:ours_clean}
\end{figure}

To build a reference line, we first conduct experiments on the clean datasets, i.e. datasets that are not attacked. Figure \ref{fig:ours_clean} demonstrates how the ratio of $R_{\tilde{A^{*}}}$ and  $R_{A}$ affects the test accurary of GCN models trained on the reconstructed adjacency matrix.

In the following experiments under the targeted attack, we do not perform hyperparameter search on $R_{\tilde{A^{*}}}$ and fix $R_{\tilde{A^{*}}}/R_{A}$ to be 20 for the reason that the number of edges added or removed is extremely limited, and the time cosuconsuming hyperparameter searching brings is unnecessary as we mainly focus on the defense performance. The results shown in Table \ref{table:cleandata} are also under this setting. Note that we do not have the performance for Jaccard defense model in Polblogs since this model requires node features but Polblogs does not provide node features. We can also observe that SVD defense model decreases the accuracy on Cora and Citeseer datasets too much. The reductions of the proposed DefenseVGAE's accuracy are whin 2\% which lays the foundation for applying it to the adversarial settings.

\begin{table}[htbp]
    \caption{The results of node classification accuracy (in percentages) on clean datasets.}
    \centering
    \begin{tabular}{p{3cm}<{\centering}p{2.5cm}<{\centering}p{2.5cm}<{\centering}p{2.5cm}<{\centering}}
        \hline
        Methods & Cora  & Citeseer & Polblogs \\
        \hline
        GCN     &   80.4  &  68.1  &  95.71  \\
        \hline
        GCN-Jaccard &  78.3  &   68.5  &  --   \\
        GCN-SVD    &   64.2  &   59.4  &   93.87  \\
        RGCN       &   79.8  &   69.0  &  95.19   \\
        Ours       &   78.6  &   67.8  &  95.40   \\
        \hline       
    \end{tabular}
    \label{table:cleandata}
\end{table}

\subsection{Results on Targeted Attacks}

In this section, we first evaluate the classification accuracy of targeted nodes against targeted adversarial attacks. We adopt Nettack as the attack method. According to the algorithm proposed in \cite{Zgner2018AdversarialAO}, there are two different ways to attack a target node : direct attack called Nettack, and influence attack called Nettack-In which attacks a node indirectly.
In our experiments we only consider attacking each target node directly, as it is a stronger attack compared to an indirect attack. Also we only consider structure attack.  We first use Nettack to generate different numbers of perturbations for the targeted nodes, where one perturbation is defined as adding or deleting one edge in the graph. Then, we retrain the GCN model for poisoning attacks. Finally, we test  whether each GCN can classify its corresponding targeted node, i.e. whether we successfully defend the attack. We vary the number of perturbations from 1 to 5. The results are  reported in Table \ref{table:nattack-performance}.

From Table \ref{table:nattack-performance}, we can  identify that the performance of all methods decays rapidly with respect to the number of perturbations, demonstrating that Nettack is a very strong attack method.  By enabling our defense approach, the accuracy can be significantly elevated. Especially we can see that DefenseVGAE is consistently outperform all baselines on Cora dataset. 

\begin{table}[ht]
    \caption{Results of different method when adopting Nettack as the attack method. Note $N$ denotes the number of perturbations.  And we do not have the performance for SVD defense model in Cora and Citeseer since it sacrifices two much accuracy on clean data.}
    \centering
    \begin{tabular}{p{2cm}<{\centering}p{2.5cm}<{\centering}p{1.2cm}<{\centering}p{1.2cm}<{\centering}p{1.2cm}<{\centering}p{1.2cm}<{\centering}p{1.2cm}<{\centering}}
        \hline
        Datasets & $N$  & 1 & 2 & 3 & 4 & 5 \\
        \hline
              & GCN     &    54.8  &  24.0  &  12.0  &  6.4   &  3.6   \\
              &Jaccard  &    68.6  &  46.8  &  32.8  &  21.6  &  15.4 \\
        Cora  & RGCN    &    55.6  &  26.2  &  12.8  &  7.0   &  4.6  \\
              & Ours    &   \textbf{81.8}   &  \textbf{62.4}  &  \textbf{45.2}   &  \textbf{30.8}   &   \textbf{19.0} \\
        \hline
                  & GCN     &   28.6 & 10.8 & 5.4 & 2.4 & 1.2 \\
                  &Jaccard  &   \textbf{79.8} & \textbf{67.0} & \textbf{54.8} & \textbf{46.6} & \textbf{39.4} \\
        Citeseer  & RGCN    &   25.0 & 11.2 & 5.6 & 1.8 & 1.4   \\
                  & Ours    &   66.0 & 47.6 & 37.8 & 27.4 & 20.6 \\
        \hline  
                  & GCN     &  89.6 & 81.6 & 69.2 & 61.4 & 59.4 \\
                  & SVD     &  \textbf{96.4} & \textbf{93.4} & \textbf{88.2} & \textbf{83.8} & \textbf{79.6} \\
        Polblogs  & RGCN    &   90.2  & 80.0 & 65.8 & 57.8 & 57.4 \\
                  & Ours    &  94.0   & 90.8 & 86.2 & 81.6 & 77.2 \\
        \hline
    \end{tabular}
    \label{table:nattack-performance}
\end{table}

\subsection{Results on Untargeted Attacks}
In this section, we continue to evaluate the overall classification accuracy of different methods against untargeted adversarial attacks.
We adopt Metattack as the attack method in the form of poisoning attack and structure attack similarly.
Note that we vary the perturbations from 1\% to 5\% with a step of 1\%. The results are demonstrated in Table \ref{table:metattack-performance}.

From Table \ref{table:metattack-performance}, we can see DefenseVGAE is consistently outperform all baselines on Citeseer dataset. On Cora dataset our proposed method outperform all baselines when Metattack perturbs 1\%, 3\% and 4\% edges.

\begin{table}[htbp]
    \caption{Results of different method when adopting Metattack as the attack method. Note that $r$ denotes perturbation rate.}
    \centering
    \begin{tabular}{p{2cm}<{\centering}p{2.5cm}<{\centering}p{1.2cm}<{\centering}p{1.2cm}<{\centering}p{1.2cm}<{\centering}p{1.2cm}<{\centering}p{1.2cm}<{\centering}}
        \hline
        Datasets & $r(\%)$  & 1 & 2 & 3 & 4 & 5 \\
        \hline
              & GCN     &    77.8   &   74.7  &  72.3   &  69.3   &   67.4  \\
              &Jaccard  &    76.8   &   76.4  &  74.5   &  72.0   &   \textbf{71.6}  \\
        Cora  & RGCN    &   77.9    &   \textbf{77.3}  &  74.6   &  73.5   &   70.0  \\
              & Ours    &   \textbf{78.4}    &   76.4  &  \textbf{75.0}   &  \textbf{73.9}   &    67.8 \\
        \hline
                  & GCN     &   65.4 & 62.0 & 56.5 & 55.7 & 53.5 \\
                  &Jaccard  &   65.6 & 63.3 & 63.0 &  62.1 & 62.1 \\
        Citeseer  & RGCN    &   61.6 & 58.9 & 56.0 &57.8  &53.3   \\
                  & Ours    &   \textbf{66.1} & \textbf{63.9} & \textbf{64.2} & \textbf{63.3} & \textbf{64.6} \\
        \hline  
                  & GCN     &  84.20 & 82.92 & 79.75 & 79.14 & 77.20 \\
                  & SVD     &  \textbf{94.07} & \textbf{94.17} & \textbf{94.27} & \textbf{94.17} & \textbf{93.46} \\
        Polblogs  & RGCN    &   84.20 & 83.64  &79.55 & 78.32  & 77.10 \\
                  & Ours    &   92.74 & 93.87  &93.66 & 93.35  & 93.25 \\
        \hline
    \end{tabular}
    \label{table:metattack-performance}
\end{table}

\section{Conclusion}
In this paper, we proposed DefenseVGAE, a novel defense strategy utilizing VGAEs  to enhance the robustness of node classification models against targeted or untargeted graph adversarial attacks. We show that our proposed method is effective against commonly considered adversarial attacks and achieve a performance close to the performance on the clean graph. Under some settings it outperforms the strong baslines. We will futher investigate using DefenseVGAE to defend against attacks perturbing node features. Moreover, the success of DefenseVGAE relies on the expressiveness and generative power of the VGAE.  However, finding better-suited prior distributions and more flexible generative models is still a challenging task and an active area of research.

%
%
%
\bibliographystyle{splncs04}
\bibliography{references}
%




\end{document}